\documentclass[a4paper,twocolumn,fleqn]{article} 

\usepackage{ist}
\usepackage{times}
\usepackage{graphicx}
\DeclareGraphicsExtensions{.ps,.pdf,.png,.tif,.jpg}
\usepackage{hyperref}
\usepackage{color}
\usepackage{comment}
\usepackage{amsmath}
\usepackage{amssymb}
\usepackage{mathtools}
\usepackage{hhline}
\usepackage{caption}
\usepackage{float}
 \newcommand{\Despace}{\vspace*{-10pt}}
 \newcommand{\despace}{\vspace*{-7.5pt}}

\title{Illumination-Invariant Image from 4-Channel Images: 
\\The Effect of Near-Infrared Data in Shadow Removal}

\author{ 
Sorour Mohajerani$^1$, 
Mark S.\ Drew$^2$, 
Parvaneh Saeedi$^1$,
\{$^1$School of Engineering Science, $^2$School of Computing Science\}, Simon Fraser University, Vancouver, BC,
Canada \{smohajer/mark/psaeedi\}@sfu.ca, 
}

\date{} 

\begin{document} 

\maketitle 

\thispagestyle{empty} 


\begin{abstract}
Removing the effect of illumination variation in images has been proved to be
beneficial in many computer vision applications such as object recognition and
semantic segmentation. Although generating illumination-invariant images has
been studied in the literature before, it has not been investigated on real
4-channel (4D) data. In this study, we examine the quality of
illumination-invariant images generated from red, green, blue, and near-infrared
(RGBN) data. Our experiments show that the near-infrared channel substantively
contributes toward removing illumination. As shown in our numerical and visual results, the illumination-invariant image obtained by RGBN data is superior compared to that obtained by RGB alone.
\end{abstract}

\section{Introduction}
\label{sec:intro}

Images captured by cameras/sensors are the product of an interaction between
illumination sources, objects' surfaces, and the image acquisition device. In
regular images, light from sun/sky/an artificial illumination source impinges
upon an object and the object's surface reflects the light, with a camera
recording the reflected energy. The captured image may include shadows, which indeed represent a change in the illumination of the scene (at least one object under two different illumination conditions). 

In one research stream, removing illumination effects from regular RGB color
images has been done via log-chromaticity with camera calibration \cite{FINLAYSON.DREW.ICCV01} and entropy minimization \cite{SHADOWS.ENTROPY.IJCV.2009}. In addition, recently, deep learning-based methods have been used to remove shadows \cite{deeplearning_shadowremoval}.

However, removing illumination has not been investigated in RGBN images. These images have an extra channel of near-infrared (N) compared to regular images, which are in the visible spectrum (RGB) only. The N channel usually covers the spectrum in the range of $800-1000 nm$. RGBN images have been widely used in many fields such as remote sensing \cite{cloudnet.sorour.igarss19} , medical imaging\cite{nir_in_biomedical}, and computer vision \cite{nir_in_computer_vision}. 

In this work, we investigate the effectiveness of the N channel in the process
of shadow removal. Our experiments show that having the N channel allows
generating a better illumination-invariant image than that obtained by visible
channels alone. As a result of this benefit, the performance of shadow removal process improves.

\section{Previous Works}
Several studies have been conducted to address the problem of shadow detection
and removal in RGB images
\cite{sorour_shadow_tip,sorour_shadow_mmsp,FINLAYSON.HORDLEY.SHADOWS.PAMI06,FINLAYSON.HORDLEY.SHADOWS.ECCV02}.
However, the number of research works which address those problems by considering
near-infrared data is limited. For instance, R{\"u}fenacht et al.\
\cite{nir_shadow_detection_tpami} have used the information in near-infrared
channel to detect shadows. The authors have noticed that many objects which are
dark in the visible image are brighter in the N image. They have used this
observation to distinguish dark objects from shadows, which are dark in both
visible and near-infrared channels. The authors in
\cite{nir_shadow_removal_icip} have utilized a near-infrared channel to extract a shadow probability map to find the location of penumbra and umbra shadows. They have, then, re-lit those areas in RGB images to match the lightness (L channel in CIELab color space) of non-shadow areas.

\section{Methodology}
The energy reflected from an object is captured by a camera to form an image. If
the object's surface is modeled as a Lambertian surface, for each pixel of this
image, the following equation obtains:
\Despace \begin{equation}
  R_{k} = \sigma \!\!  \int_{}^{} \! E(\lambda, T) S(\lambda) Q_{k}(\lambda) d\lambda , \quad k \in \{1,2,3,4 \}
\label{eq:rk}
\end{equation}
where $R_{k}$ and $\sigma$ denote the intensity of a pixel for channel $k$ of
the image and the shading coefficient, respectively. $E(\lambda)$, $S(\lambda)$,
and $ Q_{k}(\lambda)$ represent the spectral power distribution of the
illumination at wavelength $\lambda$, the surface spectral reflectance of the object's surface, and the sensitivity response of the camera, respectively. $T$ is the temperature or color of the light.

\begin{figure*}[!htbp]
\centering
\includegraphics[width=0.99\textwidth]{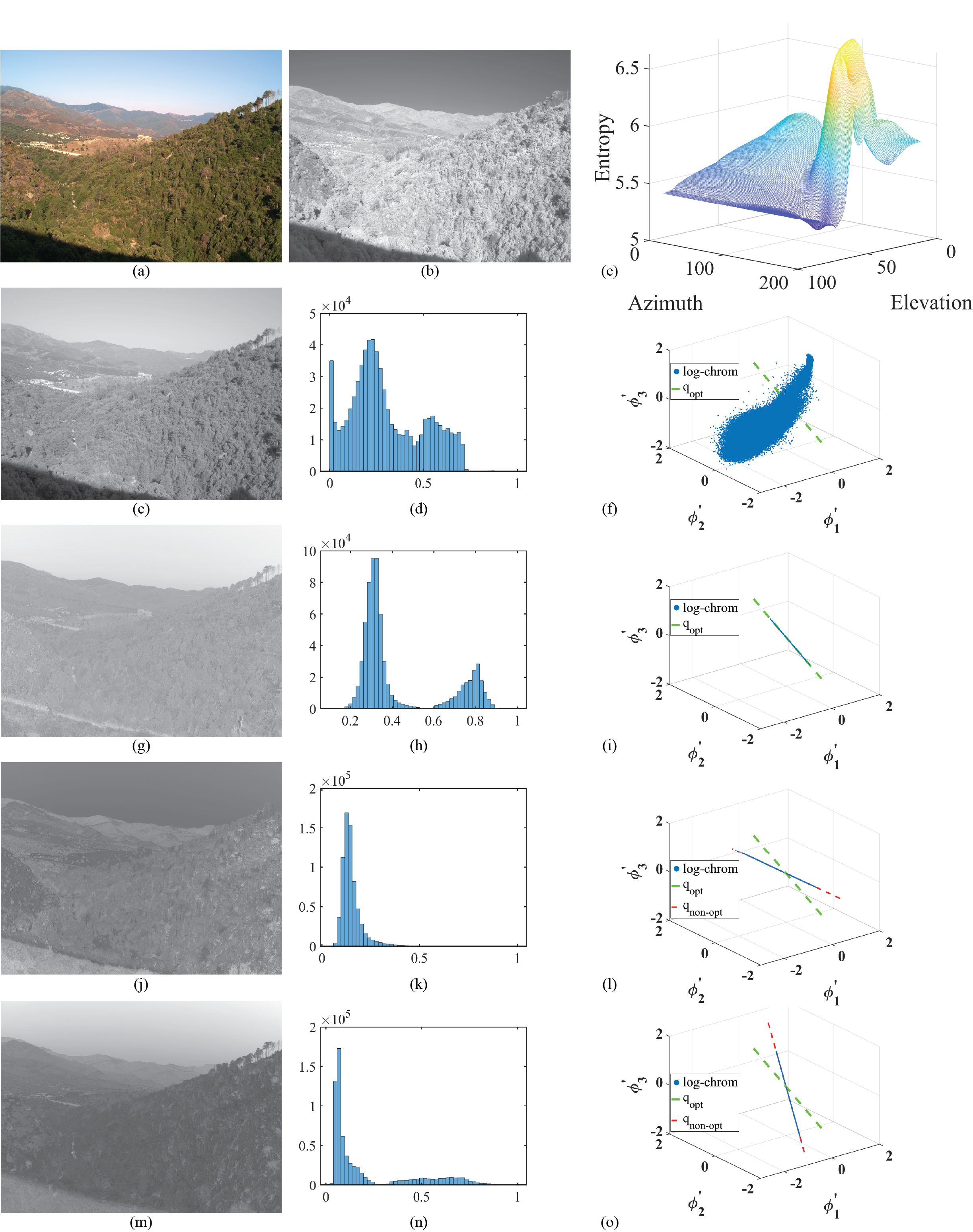}
\setlength{\abovecaptionskip}{1mm}
\caption{Illumination-invariant image formation, (a) RGB image, (b) Near-infrared channel, (c) luminance image obtained by $(R+G+B+N)/4$, (d) histogram of (c), (e) entropy variation w.r.t. $azimuth$ and $elevation$ angles, which shows the entropy of the projected data onto all of the possible $q$ vectors, (f) mapped log-chromaticities in 3D space, the vector in dashed green line depicts the optimal projection vector, (g) grayscale illumination-invariant image, (h) histogram of (g), (i) projected log-chromaticities onto the optimal projection vector, (l) and (o) projection of the data onto two non-optimal examples of $q$ (dashed red vectors), (j) and (m) the corresponding invariant images as the result of non-optimal projections, (k) and (n) histograms of (j) and (m). The entropy obtained by $q$ vectors from (i), (l), and (o) is $5.16$, $6.58$, and $5.64$, respectively.}
\label{fig:c0010}
\end{figure*}

Since the sensitivity of narrow-band cameras, $ Q_{k}$, are usually approximated with a delta function with a magnitude of $q_{k}$ at the central wavelength of each channel ($\lambda_{k}$), Eq.~(\ref{eq:rk}) is simplified to: 
\Despace \begin{equation}
\hspace{-2mm}
R_{k} = \sigma \!\!  \int_{}^{} \! E(\lambda, T) S(\lambda) q_{k} \delta(\lambda - \lambda_{k}) d\lambda = \sigma E(\lambda_{k},T) S(\lambda_{k}) q_{k}
\label{eq:rk_delta}
\end{equation}

Another approximation to have a simple model for image formation is that the illumination is restricted to the Planckian locus \cite{Wyszecki1982}. Based on Wien's approximation for temperatures between $2500K$ to $10000K$:

\Despace \begin{equation}
E(\lambda,T) \simeq I a_{1}\lambda^{-5} e^{-\frac{a_{2}}{\lambda T}}
\label{eq:ill_plnk}
\end{equation}
where $a_{1}$, $a{_2}$, and $I$ denote the Planck constant, the Boltzmann
constant, and a multiplicative coefficient $I$ to represent the  intensity of
the light. By substituting Eq.\ (\ref{eq:ill_plnk}) in Eq.\ (\ref{eq:rk_delta}),
the intensity at each pixel is as follows:

\Despace \begin{equation}
R_{k} =  \sigma I a_{1}\lambda_{k}^{-5} e^{-\frac{a_{2}}{\lambda_{k} T}} S(\lambda_{k}) q_{k} 
\label{eq:rk_simplest}
\end{equation}
Each image pixel, with four values corresponding to the four channels of RGBN,
satisfies Eq.~(\ref{eq:rk_simplest}). We can eliminate the effect of shading and
light intensity in the image by dividing pixel values by one of the R, G, B, N
values. For instance by dividing by the G channel: 
\Despace \begin{equation}
\frac{R_{k}}{R_{G} } =  (\frac{\lambda_{k}}{\lambda_{G}})^{-5} e^{\frac {-a_{2}}{T}(\frac{1}{\lambda_{k}} - \frac{1}{\lambda_{G}})} \frac{S(\lambda_{k})q_{k}}{S(\lambda_{G})q_{G}}
\label{eq:chrom_r/g}
\end{equation}

By taking logarithms on both sides of Eq.~(\ref{eq:chrom_r/g}), we form the band-ratio chromaticities \cite{FINLAYSON.HORDLEY.SHADOWS.PAMI06}:
\Despace \begin{equation}
    \Psi_{k} \! \equiv  log(\frac{R_{k}}{R_{G}}) \! = \! \frac {-a_{2}}{T}(\frac{1}{\lambda_{k}} \! - \! \frac{1}{\lambda_{G}}) + log(\frac{S(\lambda_{k})q_{k}\lambda_{k}^{-5}}{S(\lambda_{G})q_{G}\lambda_{G}^{-5}})
    \label{eq:logchrom_r/g}
\end{equation}

We can form another version of chromaticity to have even a simpler and more convenient equation than Eq.~(\ref{eq:logchrom_r/g}). Instead of dividing pixel values by one of the channels, we divide them by the geometric mean of the pixels. This approach not only avoids bias towards one of the channels but also leads to a format in which log-chromaticities can be further simplified. 

The geometric mean of an image at each pixel is defined as:
\Despace \Despace \begin{equation}
    R_{m} =  \bigg ( \prod_{k=1}^{N_{k}} R_{k} \bigg ) ^{\frac{1}{N_{k}}} 
    \label{eq:geogmean}
\end{equation}
where $N_{k} \! = \!4$ for RGBN images. Forming a log-chromaticity space by
dividing pixel values by the geometric mean and taking logs, we will have the following equation:

\Despace \despace \begin{equation}
\hspace{-2mm}
    \Phi_{k} \! \equiv  log(\frac{R_{k}}{R_{m}}) \! = \! \frac {-a_{2}}{T}(\frac{1}{\lambda_{k}} \! - \! \frac{1}{\lambda_{m}}) + log(\frac{S(\lambda_{k})q_{k}\lambda_{k}^{-5}}{S(\lambda_{m})q_{m}\lambda_{m}^{-5}})
    \label{eq:logchrom}
\end{equation}
where $\lambda_{m}$ and $q_{m}$ denote the geometric mean of central wavelengths and geometric mean of camera sensitivity magnitudes, respectively. This equation can be rewritten as follows:

\Despace \despace \begin{alignat}{2}
   & \Phi_{k} = \! \frac {1}{T}( e_{k} - e_{m}) + log(\frac{s_{k}}{s_{m}}), \\ 
   & \nonumber e_{k} = \frac{-a_{2}}{\lambda_{k}},  \quad  e_{m} = \frac{-a_{2}}{\lambda_{m}}, \quad s_{k} = S(\lambda_{k})q_{k}\lambda_{k}^{-5}, \\
   & \nonumber s_{m} = S(\lambda_{m})q_{m}\lambda_{m}^{-5}
    \label{eq:logchrom_simpl}
\end{alignat}
This formulation shows these important points: (1) log-chromaticty values of an
image at a certain type of surface indeed live on a line (in a 4D space)
parametrized by $T$, with slope $e_{k} - e_{m}$; (2) all pixels belonging to one surface, under multiple illumination colors (multiple $T$), are located on that line; (3) the offset of the line ($log(s_{k} / s_{m})$) is independent of the illumination and represents only surface characteristics.

Having used geometric means for getting chromaticities, the log-chromaticities
satisfy the following criteria at each pixel in 4D space: $\Phi_{1}+ \Phi_{2} +
\Phi_{3} + \Phi_{4} =0$. This, in fact, is the equation of a subspace in 4D
space orthogonal to the normal vector $ u = [1, 1 ,1 ,1]^{T}/\sqrt4 $. To get
the equivalent location of log-chromaticity values in the 3D space, we project
them onto the subspace orthogonal to $u$. This leads to a dimension reduction
from 4D to 3D, ($\Phi_{k} \in \mathbb{R} ^{4} \to \Phi'_{l} \in \mathbb{R} ^{3} $). 

Having mapped log-chromaticities in 3D, they can be used to remove the effect of
illumination in the image. This is done via projecting $\Phi'_{l}$ onto a vector
orthogonal to the direction of $v = e_{k} - e_{m}$. The result of this
projection is a grayscale image in which the same surfaces under two
illumination would have the same values values as each other. That is why this
image is called the grayscale illumination-invariant, or intrinsic, image. In other words, shadows will be removed in the invariant image (since shadow areas in an image are caused by different illumination conditions from non-shadow parts):

\Despace \despace \begin{equation}
    I = \Phi'_{l} (v^{\perp})^{T}
    \label{eq:projection}
\end{equation}
where $I$ and $v^{\perp}$ denote the projected image and the vector orthogonal to illumination change, respectively.

\begin{figure}[t]
\centering
\includegraphics[width=0.48\textwidth]{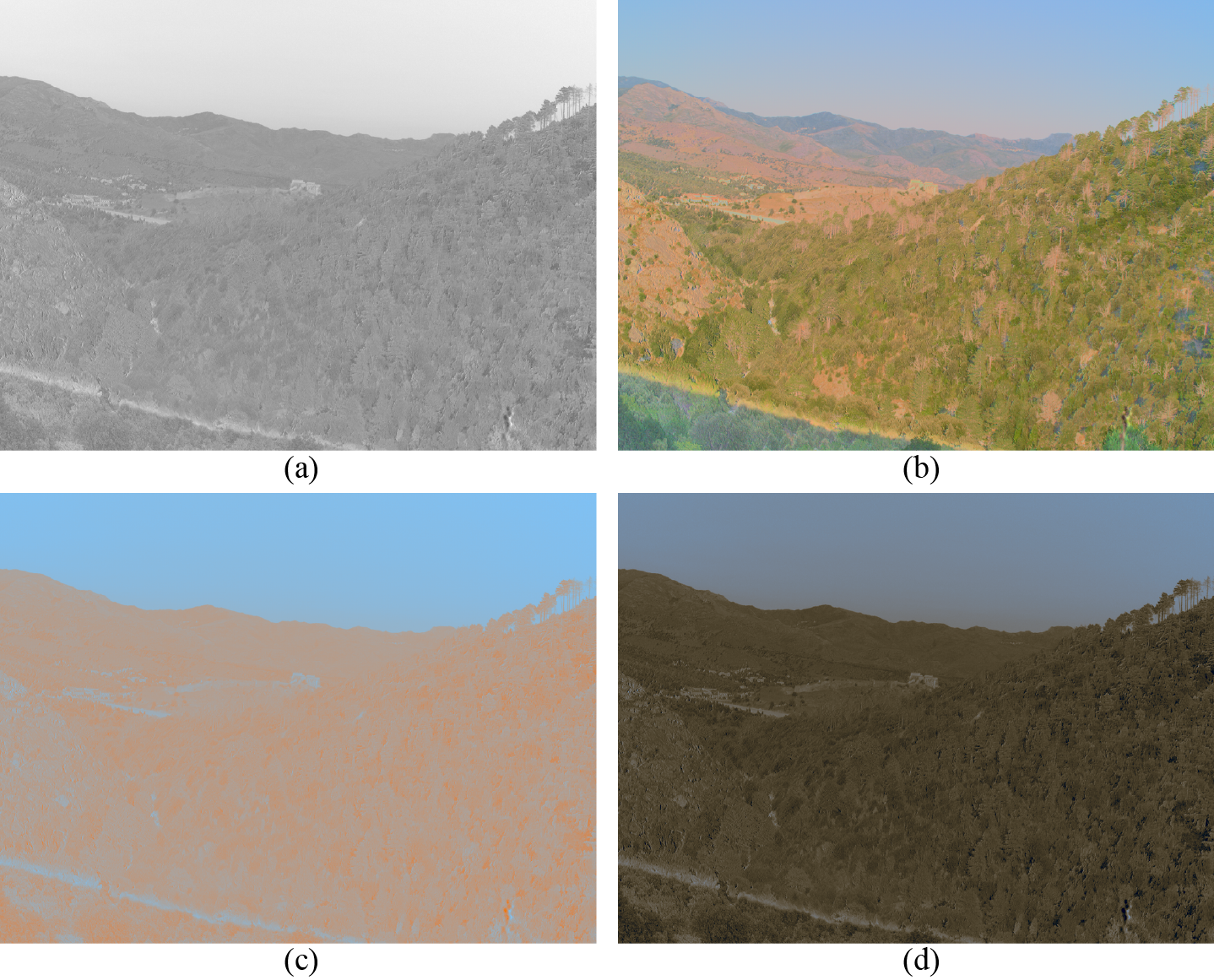}
\setlength{\abovecaptionskip}{-1mm}
\caption{Invariant RGB image formation, (a) illumination-invariant image, (b) first three channels of the L1-chromaticity, (c) invariant RGB image obtained by projection (Eq.~(\ref{eq:3d_chrom_proj}), (d) invariant RGB image obtained by multiplication ( Eq.~(\ref{eq:3d_chrom_l1})).}\vspace{5mm}
\label{fig:c0010_chrom}
\end{figure}

Since the exact direction of $v$ is unknown, following \cite{FINLAYSON.DREW.SHADOWS.ECCV04}, we can address this problem by searching over all possible projection vectors and find out which of those leads to \textit{minimum entropy} in the projected image:

\Despace \despace \begin{equation}
    q_{opt} = \operatorname*{argmin}_q \bigg (- \sum_{i=0}^{1} p_{i}(\Phi'_{l} q^{T}) log(p_{i}(\Phi'_{l} q^{T})) \bigg)
    \label{entropymin}
\end{equation}
where $q$ and $q_{opt}$ represent projection vector and the optimal projection vector, respectively. $p_{i}$ denotes the probability of having pixel values equal to $i$ in the projected image. To attenuate noise, entropy is calculated using only $90\%$ of the data---excluding first and last five percentiles of the data. In addition, the number of possible bins (discrete values of $i$) is limited by Scott's Rule \cite{SCOTT.MULTIVARIATE.92}.

Re-identifying $q$ in polar coordinates rather than Cartesian limits the search space for entropy minimization. Therefore, instead of finding three unknown parameters in $q= [q_{1}, q_{2}, q_{3}]^{T}$, we search for two unknown angles of $elevation$ and $azimuth$ in $q=[\cos{el} \cos{az}, \cos{el} \sin{az}, \sin{el}]^{T}$ (radius is set to $1$).

After finding the proper vector, the grayscale invariant image is obtained by:
\Despace \begin{equation}
  \quad I_{inv} = exp(\Phi'_{l} q_{opt}^{T})
    \label{eq:grayinvimage}
\end{equation}

All of the described steps are illustrated in Fig.\ (\ref{fig:c0010}). To get a shadow-free version of the image in 3D space, we need to project $\Phi'_{l}$ onto the found vector $q_{opt}$ while preserving the final coordinates in 3D. This is done via a $3\times3$ projector ($P_{q_{opt}}$):

\Despace \begin{equation}
   \widetilde {\Phi}{'_{l}} = \Phi'_{l} P_{q_{opt}}, \quad  \widetilde{I}={exp(\widetilde {\Phi}{'_{l}})}
   \label{eq:3d_chrom_proj}
\end{equation}
$\widetilde{I}$ represents a shadow-free RGB image (illustrated in Fig. \ref{fig:c0010_chrom}(c)). We have found that with a simple modification we can get a visually nicer shadow-free RGB image. First, the following equation is solved to find a $3\times3$ transformation matrix $M$ which maps the obtained $\widetilde{I}$ to the first three channels of L1-chromaticity of the original image ($\rho = \{R,G,B\}/(R\!+\!B\!+\!G\!+\!N)$): $\rho\approx  \widetilde{I} M$. Then, we calculate the approximated image and multiply each channel of it by the illumination-invariant image. This way, the values are forced to mimic/copy the patterns in invariant image, which leads to a better recovered shadow-free RGB image:  
\Despace \begin{equation}
    \widetilde{I}_{aprx} = \widetilde{I}M, \quad I_{rec}= I_{inv}\widetilde{I}_{aprx}
    \label{eq:3d_chrom_l1}
\end{equation}
where $I_{rec}$ represents the final shadow-free RGB image (Fig. \ref{fig:c0010_chrom}(d)).


\section{Experimental Results}
\label{sec:res}

To quantify the contribution of the N channel to shadow removal in an image, we have compared the illumination-invariant image obtained by RGBN to that of obtained by RGB. If a shadow area is perfectly removed, regions inside that area should be similar to their adjacent non-shadow regions. In other words, two sub-regions belonging to a surface---under two different illuminations---should share similar qualities.

We have selected $6$ images from the RGB-NIR public dataset \cite{rgbn_dataset} to conduct our experiments. The images of this dataset have been captured by Nikon D90 and Canon T1i cameras. For each scene, there are two images: (1) an RGB image captured by B+W $486$ (visible) filter, (2) a grayscale near-infrared image captured by a $093$ filter. 

We have utilized the root mean squared error (RMSE) to measure the similarity of two adjacent sub-regions:

\Despace \begin{equation}
    D = I^{sh}_{inv}\!  - I^{nonsh}_{inv},  \quad  RMSE \! = \! \sqrt{(1\mathbin{/}M) \sum\nolimits_{i=1}^{M} D^{2}_{i}} \despace
    \label{eq:metric}
\end{equation}
where $RMSE$ denotes the error between two same size sub-regions in illumination-invariant image: one inside shadow ($I^{sh}_{inv}$) and the other one outside shadow ($I^{nonsh}_{inv}$). $M$ represents the total number of pixels existing in a sub-region.

\begin{figure}[t]
\centering
\includegraphics[width=0.45\textwidth]{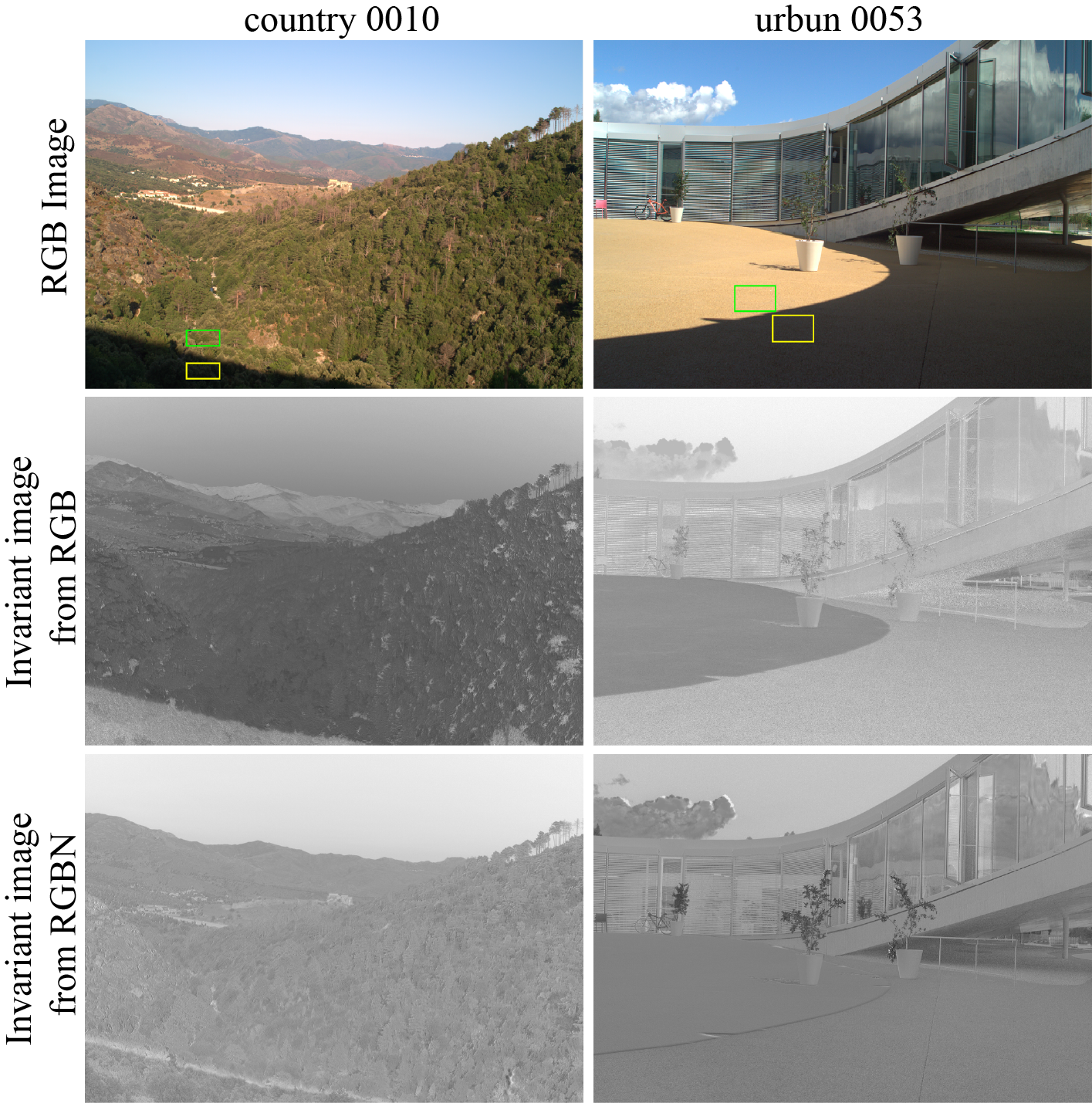}
\setlength{\abovecaptionskip}{1mm}
\caption{Some examples of illumination-invariant images. The sub-regions selected to calculate error are displayed in yellow and green boxes.} \vspace{3mm}
\label{fig:visual}
\end{figure}

The $RMSE$ is calculated for the illumination-invariant image of RGBN and RGB.
Table \ref{Tab:numerical} shows the numerical results. According to this table,
in all of the tested images the error between a shadow region and its
neighbourhood non-shadow one is smaller in the 4D invariant image than that from a 3D image. Fig. \ref{fig:visual} illustrates some of the visual results.  In this figure, clearly, the contrast between shadow and non-shadow areas are less visible in the invariant images obtained by 4D data (bottom row) than the ones obtained by 3D data (middle row). In addition, the edge artifact (between shadow and non-shadow area) formed in an invariant image is less observable in the image obtained from 4D than the one obtained form 3D. The reason for having such an artifact is that an optimal projection vector is selected in order to minimize the entropy in the entire image. However, this very vector might not be a proper value for the regions in immediate adjacency to shadow areas. This suggests that further considerations in the optimization process are required for removing this artifact in the regions close to the boundary between shadow and non-shadow areas. 
\vspace{5mm}

{\renewcommand{\tabcolsep}{5pt}
\begin{table}[!h]
\begin{minipage}{0.48\textwidth}
\centering
\small
\caption{Quantitative evaluation of shadow removal  in illumination-invariant images (in \%). Bold numbers indicate lower $RMSE$.
\label{Tab:numerical}} 
\renewcommand{\arraystretch}{1.2}
\begin{tabular}{|c|c|c|}
\hline
\centering \textbf{Image Name}   & \textbf{RMSE of RGBN}  & \textbf{RMSE of RGB} \\
\hhline{|=|=|=|}
country 0010 & \textbf{5.078} & 18.055 \\ \hline
field 0052   & \textbf{0.516} & 0.590 \\ \hline
mountain 0026  & \textbf{3.755} & 10.497\\ \hline
old building 0004 & \textbf{0.904} & 1.206\\ \hline
street 0001  & \textbf{0.334} & 0.457\\ \hline
urban 0053  & \textbf{1.572} & 20.953\\ \hline
\end{tabular}
\end{minipage}
\end{table}}


\section{Summary and Conclusions}
The effectiveness of using a near-infrared channel in the shadow removal process has been investigated in this paper. Illumination-invariant images obtained by the 4D RGBN inputs numerically and visually deliver better quality than that of RGB inputs. This indicates that having more spectral data (such as near-infrared) helps to remove the illumination effects in a scene. 

\section{Future Works}
Since removing shadows is an important pre-processing step in many remote sensing applications, employing the proposed method on multispectral satellite images could be a future direction. Additionally, a more sophisticated optimization algorithm for finding the optimal projection vector could be utilized. This could be beneficial for removing the edge artifacts observed in Fig. \ref{fig:visual}.




\small
\bibliographystyle{IEEEbib} 
\bibliography{refs}
\normalsize
\begin{biography}

Sorour Mohajerani received the B.Sc. and M.Sc. degrees in electrical engineering from Tehran Polytechnic and Iran University of Science and Technology. In 2017, she joined the School of Engineering Science, Simon Fraser University, Canada, as a Ph.D. student. Her current research interests include computer vision, remote sensing, and deep learning.

Mark S. Drew is a Professor in the School of Computing Science at Simon
Fraser University in Vancouver. His background is in
Engineering Science, Mathematics, and Physics.  His interests lie in the
fields of image processing, color, computer vision, computer graphics,
and multimedia. He has published over 160 refereed papers.

Parvaneh Saeedi (Ph.D., P.Eng.) is an Associate Professor at the School of Engineering Science at Simon Fraser University, Burnaby, BC, Canada. Her research interests include machine learning, image processing, pattern recognition, computer vision, and artificial intelligence.
\end{biography}

\end{document}